\documentclass[11pt,a4paper]{article}

\usepackage[preprint]{acl}      
\usepackage{times}              
\usepackage{latexsym}
\usepackage[T1]{fontenc}
\usepackage[utf8]{inputenc}

\usepackage{newunicodechar}
\newunicodechar{·}{\textperiodcentered}
\newunicodechar{—}{---}
\newunicodechar{–}{--}
\newunicodechar{…}{\ldots}
\newunicodechar{“}{``}
\newunicodechar{”}{''}
\newunicodechar{‘}{`}
\newunicodechar{’}{'}
\newunicodechar{→}{$\rightarrow$}
\newunicodechar{≤}{$\leq$}
\newunicodechar{≥}{$\geq$}
\newunicodechar{≈}{$\approx$}
\newunicodechar{κ}{$\kappa$}
\newunicodechar{σ}{$\sigma$}
\newunicodechar{τ}{$\tau$}
\newunicodechar{Φ}{$\Phi$}
\newunicodechar{Ψ}{$\Psi$}
\newunicodechar{∅}{$\emptyset$}
\newunicodechar{×}{$\times$}
\newunicodechar{∥}{$\parallel$}
\newunicodechar{§}{\S}

\usepackage{microtype}
\usepackage{graphicx}
\usepackage{amsmath, amssymb, amsthm}
\usepackage{booktabs}
\usepackage{multirow}
\usepackage{xcolor}
\usepackage{listings}
\usepackage{tikz}

\usepackage{titlesec}
\titlespacing*{\section}{0pt}{5pt plus 1pt minus 1pt}{1pt plus 1pt}
\titlespacing*{\subsection}{0pt}{3pt plus 1pt minus 1pt}{0pt}
\titlespacing*{\subsubsection}{0pt}{2pt plus 1pt minus 1pt}{0pt}
\titlespacing*{\paragraph}{0pt}{1pt plus 1pt minus 1pt}{0.5em}
\usepackage{enumitem}
\setlist{topsep=1pt,itemsep=0pt,parsep=0pt,partopsep=0pt}
\usepackage[font=small,labelfont=bf,skip=3pt]{caption}

\usepackage[]{natbib}

\hypersetup{colorlinks=true,
            linkcolor=black,
            citecolor=black,
            urlcolor=black!80!blue}
\usepackage{cleveref}

\newcommand{\W}{\mathcal{W}}
\newcommand{\slive}{s_{\text{live}}}
\newcommand{\sstatic}{s_{\text{static}}}
\newcommand{\LLM}{\mathcal{L}}
\newcommand{\sic}{\slive^{\text{ic}}}
\newcommand{\sfs}{\slive^{\text{fs}}}
\newcommand{\ssh}{\slive^{\text{sh}}}

\title{State-Grounded Conditioning:\\
       Wrapping User-Facing LLM Agents\\
       Where Direction Depends on Live State}

\author{
  Qi Liu \quad
  Xiaoyang Yuan \quad
  Yubin Ruan \quad
  Zhuomeng Zhang \quad
  Wenjin Wang \\
  Di Wu \quad
  Mingye Xu \quad
  Xinyi Mou \quad
  Xingxi Yin \quad
  Ke Feng \quad
  Zixun Sun \\[2pt]
  Tencent
}

\begin{document}
\maketitle

\begin{abstract}
We introduce \textbf{State-Grounded Conditioning (SGC)}, a
design principle for user-facing LLM agents that must
condition on live user state (game state, session history,
live inventory), and a distinct failure class we call
\textbf{direction drift}: task-complete responses whose chosen
direction misaligns with the current state. SGC externalises
state-dependent control into rule kernels over structured inputs
and three primary state slices, via \emph{Perception},
\emph{Grounding}, and \emph{Interaction} wrappers with explicit
conditioning dependencies. We evaluate SGC on a 200-session
anonymised benchmark ($\approx\!1{,}000$ assistant model
turns) from an in-game conversational coaching agent that
guides players through consecutive competitive matches,
reporting mean first-token latency and five human-annotated
dialogue-quality metrics that jointly cover factual grounding
and coach-like guidance progression. The Perception wrapper
holds mean first-token latency at $1.5$\,s (vs.\ $6.1$\,s for
PE-Agent~\citep{wang2023planandsolve} inside a production
tool-use harness~\citep{xu2026lifeharness}); enabling all three
wrappers lifts turn-level grounded accuracy from $61.1\%/69.8\%$
(Prompting / PE-Agent) to $96.7\%$ and session-level grounded
accuracy from $20.0\%/26.5\%$ to $83.5\%$; session-level
grounding-failure incidents drop by $\approx\!78\%$ relative
to the strongest baseline. A cumulative ablation shows
complementary incremental gains as the wrappers are added.
These results inform \emph{approximate} state-slice orthogonality,
without establishing independent per-wrapper effects.
\end{abstract}

\section{Introduction}
\label{sec:intro}

\paragraph{From answering to coaching.}
In in-game dialogue, prompt-based and agent-harness solutions
can answer questions but struggle to act as coaches: replies
are shallow, responses arrive late, phrasing lacks coherence,
and they fail to keep players progressing through their real
in-match problems.

\paragraph{Live user-facing agents.}
Deployed LLM agents now power live customer-facing products
such as Airbnb's Agent-in-the-Loop customer support
\citep{zhao2025aitl} and real-time in-game conversational
coaching agents. In these settings, the user's \emph{live
state} (recent utterances, current inventory, session
history) becomes an inference-time input the agent must
condition on within tight latency budgets.

\paragraph{A shared gap: direction drift.}
Existing production harness work
\citep{xu2026lifeharness,park2025workflow} standardises
\emph{how} an agent is orchestrated (planning, tool use,
memory), but leaves \emph{what live state each decision is
conditioned on} implicit inside a single autoregressive pass.
Under this common design, we observe a class of failures we
call \textbf{direction drift}: outputs are grammatical,
on-topic, and even ``task-complete'' by conventional metrics,
yet their \emph{chosen direction} misaligns with the user's
current state. In an online-deployed LLM-based
conversational coaching agent that guides players in
competitive matches (Section~\ref{sec:benchmark}), four
manifestations shaped our design:

\begin{itemize}[leftmargin=1.4em,itemsep=0pt,topsep=2pt]
  \item \textbf{(A) Latency drift.} Under 2-second first-token
        budgets, the planner commits to a tool call before
        live game state is loaded.
  \item \textbf{(B) Slot drift.} The reply LLM hallucinates
        entities the user does not own, or apologises for
        entities the user does own.
  \item \textbf{(C) Fatigue drift.} Across turns, the model
        repeatedly steers toward the same recommendation.
  \item \textbf{(D) Coherence drift.} A reply makes incompatible
        judgements about the same constraint (``all required
        pets are available'' alongside ``required Pet~A is
        missing'').
\end{itemize}

\paragraph{Cause and proposal.}
We attribute direction drift to a single design cause: a
monolithic LLM is asked to \emph{simultaneously} carry
state-dependent decisions (which tools to freeze, which slots
are factually valid, whether to repeat a past recommendation)
inside its implicit reasoning and to render the surface reply.
We propose \textbf{State-Grounded Conditioning (SGC)}: externalise
these control decisions into rules over structured interpretations
and live state, and inject the results into generation context.
Rule kernels are deterministic for fixed inputs and configuration;
LLM interpretation and wording need not be. Three wrappers have
separate primary state slices and explicit conditioning
interfaces; orthogonality concerns their dominant effects,
not independence:

\begin{itemize}[leftmargin=1.4em,itemsep=0pt,topsep=2pt]
  \item $\W_1$ \textbf{Perception Wrapper} (intent-context
        slice, targets drift~(A))---single-pass 4-in-1 streaming
        decoder with state-aware intent-to-tool freezing.
  \item $\W_2$ \textbf{Grounding Wrapper} (fact-slots slice,
        targets drifts~(B) and~(D))---independent slot extractor with
        deterministic conflict diff and paragraph mutex.
  \item $\W_3$ \textbf{Interaction Wrapper} (session-history
        slice, targets drift~(C))---cross-turn pure-function scheduler
        with an injectable random seed (reproducible replay).
\end{itemize}

Figure~\ref{fig:sgc-arch} illustrates the pipeline.

\paragraph{Evidence and contributions.}
A cumulative ablation on the deployment-derived benchmark
shows complementary gains as wrappers are added
(Section~\ref{sec:benchmark}). It characterises approximate
state-slice orthogonality, not independent module effects.
The Perception wrapper ($\W_1$) holds mean first-token latency
at $1.5$\,s (vs.\ $6.1$\,s for PE-Agent inside a production
tool-use harness and $2.3$\,s for Prompting); enabling all
three wrappers lifts turn-level grounded accuracy (TGA) from
$61.1\%/69.8\%$ to $96.7\%$ and session-level grounded
accuracy (SGA) from $20.0\%/26.5\%$ to $83.5\%$ on the
200-session benchmark; session-level grounding-failure
incidents drop by $\approx\!78\%$ relative to the strongest
baseline (Section~\ref{sec:benchmark}).
Our contributions are:

\begin{enumerate}[leftmargin=1.6em,itemsep=1pt,topsep=2pt,label=(\roman*)]
  \item We formalise \textbf{direction drift} as a failure
        class distinct from task-completion failures targeted
        by generic harness work
        \citep{xu2026lifeharness, park2025workflow}.
  \item We propose \textbf{SGC} as three approximately
        orthogonal wrappers over three disjoint state slices,
        characterised by the cumulative comparisons above.
  \item We provide \textbf{analytical transfer} to code
        completion, e-commerce recommendation explanation, and
        voice-first assistants, showing each wrapper's
        applicability conditions and composability.
\end{enumerate}

\section{Related Work}
\label{sec:related}

\textbf{Generic agent harness.} Recent work formalises the
runtime ``harness'' wrapping LLM agents in production. At the
loop level, ReAct \citep{yao2023react} and Plan-and-Execute
\citep{wang2023planandsolve, xu2023rewoo} provide reason--act
and planner--executor patterns; at the runtime level,
LIFE-HARNESS \citep{xu2026lifeharness} adapts harness
parameters across 18 backbones and 7 environments
($\tau$-bench~\citep{yao2024taubench}, $\tau^2$-bench
\citep{barres2025tau2bench}, AgentBench
\citep{liu2024agentbench}, WebArena
\citep{zhou2024webarena}); Workflow Graphs
\citep{park2025workflow} structure agents as deterministic
graphs; the harness survey \citep{li2026harnesssurvey}
catalogues these efforts; LongMemEval
\citep{wu2024longmemeval} benchmarks long-horizon
session-history conditioning, with which $\W_3$ is
architecturally aligned. \emph{Our position}: SGC operates
\emph{above} generic harness. Where harness work optimises
\emph{whether the task completes}, SGC targets \emph{whether
the chosen direction is correct given live state}; the two
are composable. Section~\ref{sec:benchmark} instantiates the
family as \textbf{PE-Agent}.

\textbf{Fast-slow and structured/constrained generation.}
Talker-Reasoner \citep{christakopoulou2024talker} splits a
dual-system agent; Medusa \citep{zagyva2025medusa}
parallelises decoding; SLOT \citep{shen2025slot}, GCD
\citep{raspanti2025gcd}, and FACTS Grounding
\citep{jacovi2024facts} shape or evaluate output structure and
factuality. \emph{Our position}: $\W_1$ combines intent
recognition and planning in one stream, dispatching tools early
so live results can inform the separate reply generation.
$\W_2$ computes conflict diffs and paragraph eligibility before
a scheduler selects compatible candidates. It targets opposite
judgements about the same constraint, not praise--apology
co-occurrence in general or guaranteed semantic correctness.

\textbf{Retrieval and production deployment.} Agentic RAG
\citep{singh2025agenticrag} and MemGPT
\citep{packer2023memgpt} enrich generation via retrieved text
or managed memory; Agent-in-the-Loop \citep{zhao2025aitl}
closes a post-deployment data flywheel; \citet{kweon2025vta}
report a large virtual teaching assistant and the LLM Game
Agents survey \citep{hu2024gamesurvey} covers autonomous NPC
play. \emph{Our position}: RAG provides retrieved evidence;
SGC derives structured control variables from parsed inputs
and live state. Deterministic rules do not ensure input
correctness. AITL improves agents \emph{after} deployment; SGC
concerns \emph{how they are designed} before deployment.
Game-agent work has focused on autonomous NPC play; we address
\emph{real-time coaching for a human user}, where the agent
must serve rather than replace. Table~\ref{tab:baseline-comparison}
summarises how these lines relate to SGC.


\section{State-Grounded Conditioning: The Design Principle}
\label{sec:sgc}

\subsection{Problem Formulation}
\label{sec:sgc-formulation}

Decompose the LLM's inference-time context into \textbf{static
context} $\sstatic$ (prompts, system messages, task
specification---invariant within a session) and \textbf{live
state} $\slive$ (dynamic per-user, per-session observations:
game state, session history, inventory). We define
\textbf{direction drift} as the event where a task-complete
response misaligns with the correct direction implied by
$\slive$:
\begin{equation}
\label{eq:drift}
\begin{aligned}
\mathrm{drift}(y, \slive) \;=\; &
  \mathbb{I}[\, y \text{ task-complete} \,] \\
  & {}\cdot \mathbb{I}[\, y \text{ mis-directs } \slive \,],
\end{aligned}
\end{equation}
where $y$ \emph{mis-directs} $\slive$ if human adjudication
under the protocol of Appendix~\ref{app:benchmark-protocol}
labels $y$'s chosen direction as inconsistent with the
reference direction $y^{\star}(\slive)$ derived from live
state. The reference $y^{\star}(\slive)$ is an operational
ground truth from two-annotator agreement adjudicated by a
third (inter-annotator agreement statistics reported in
Limitations~L6 and Appendix~\ref{app:benchmark-protocol}).
Direction drift is distinct from hallucination (factuality of
\emph{static} content) and task-failure (whether the surface
task completes). A user owns Pet~A (which learned Skill~$X$ via
a rare unlock) and Pet~B (unevolved). She asks: \emph{``Which
of my pets counters Pet~Y?''} A reply \emph{``Recommend Pet~A
with Skill~$X$''} is \textbf{not a hallucination}---every
atomic fact is true---yet \textbf{it is direction drift}:
Pet~A's Skill~$X$ is \emph{not unlocked for this user}, and
the correct direction is Pet~B evolved. Hallucination's
reference is user-independent; direction drift's is per-user,
per-moment. Prompt engineering cannot condition on $\slive$ at
inference time (prompts are authored offline while $\slive$
arrives per-turn); RAG closes the \emph{static} knowledge gap,
not the \emph{dynamic} direction gap. A row-by-row comparison
against five neighbouring failure notions
(hallucination~\citep{ji2023hallucinationsurvey},
personalization failure, context-adherence, commonsense
constraint violation, classical DST
error~\citep{henderson2015dst}) is given in
Appendix~\ref{app:contracts}.

\subsection{The SGC Principle}

We propose \textbf{State-Grounded Conditioning}: externalise
state-dependent control decisions into explicit rules over
structured interpretations and live state, then inject their
outputs as conditioning variables. Formally:

\begin{equation}
\label{eq:sgc}
\begin{aligned}
c_1 &= \W_1(x, \sic), \\
c_2 &= \W_2(x, \sfs, c_1), \\
c_3 &= \W_3(x, \ssh, c_1), \\
y   &= \LLM(x,\, \sstatic,\, c_1,\, c_2,\, c_3).
\end{aligned}
\end{equation}

Determinism applies to the rule kernels given fixed parsed
inputs, state, and configuration (Equation~\ref{eq:sgc};
including the scheduler's
random source), not to upstream LLM interpretation or final
wording. The LLM retains natural-language generation, conditioned
on these explicit control decisions.

\subsection{Approximate State-Slice Orthogonality}
\label{sec:orthogonality}

We assign three primary slices of $\slive$: $\sic$
(\textbf{intent-context})---utterance intent, snapshot-time
environment, derivable entity links; $\sfs$
(\textbf{fact-slots})---demand and satisfiable slots from
inventory/catalog/retrieval; $\ssh$ (\textbf{session-history})---
metadata from the prior $K{=}20$ turns. This partitions source
fields, not all accessible information: $c_1$ conditions both
$\W_2$ and $\W_3$. Thus separate responsibilities do not imply
computational or statistical independence.

\emph{Approximate orthogonality} denotes differentiated primary
roles with cross-slice effects. The cumulative comparisons in
Section~\ref{sec:ablation} measure incremental gains conditional
on preceding wrappers, including dependency propagation; they
do not establish isolated effects or invariance to addition
order. State-field separation alone is not an empirical proof
(Appendix~\ref{app:contracts}). SGC is composable with generic harness
\citep{xu2026lifeharness, park2025workflow}, RAG
\citep{singh2025agenticrag}, and output-format constraints
\citep{shen2025slot, raspanti2025gcd}---all coexist in our
system; SGC contributes a fourth axis: direction alignment via
state-coupled deterministic conditioning.

\begin{figure}[t]
  \centering
  \includegraphics[width=0.95\columnwidth]{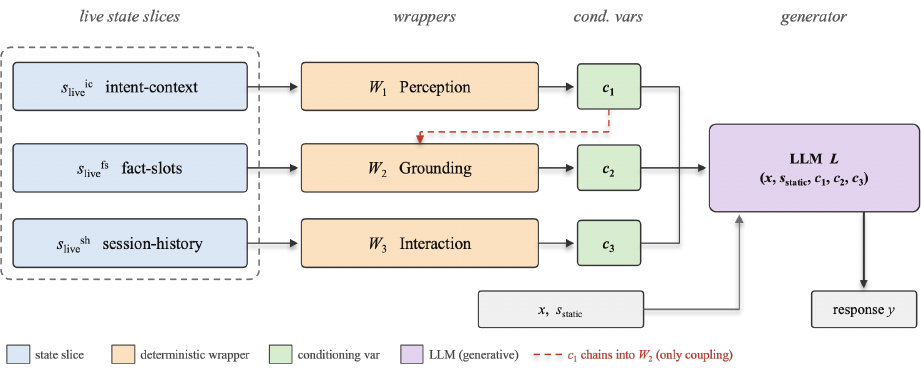}
  \caption{SGC: primary state slices and explicit conditioning
    dependencies. Dashed arrow carries $c_1$ into $\W_2$; separate
    responsibilities do not imply independent computations.
    Reply LLM $\LLM$ additionally conditions on shared input $x$
    and static state $\sstatic$.}
  \label{fig:sgc-arch}
\end{figure}

\section{Perception Wrapper: State-Aware Intent-to-Tool Freezing}
\label{sec:perception}

\paragraph{Failure mode.}
In a serial agent pipeline, intent classification and planning
consume time needed for tool execution and live-state retrieval.
Under a tight response budget, the reply may
commit to a direction before current results arrive. This
reliance on stale state constitutes \textbf{latency-induced
direction drift}.

\paragraph{Mechanism.}
$\W_1$ combines intent recognition and planning in one streaming
LLM call; final reply generation remains separate. The call
emits four segments in a fixed, intent-first order:

\begin{enumerate}[leftmargin=1.4em,itemsep=0pt,topsep=2pt]
  \item \texttt{<INTENT>}---first-level intent.
  \item \texttt{<SUB\_INTENT>}---second-level intent.
  \item \texttt{<NARRATION>}---prefatory phrase, streamed to TTS
        as tokens arrive.
  \item \texttt{<PLAN>}---fallback tool calls.
\end{enumerate}

An incremental state-machine parser triggers \textbf{early
decision} when \texttt{<SUB\_INTENT>} closes:

\begin{equation}
\label{eq:phi}
\hat{T} = \Phi\bigl(\text{intent},\, \text{sub\_intent},\, \sic\bigr).
\end{equation}

$\Phi$ (Equation~\ref{eq:phi}) is a \textbf{partial function}
over $\sim$50 registered
(intent, sub-intent) pairs. On a lookup hit with the required
current-state snapshot, $\hat{T}$ is frozen and dispatched
\textbf{without waiting for narration or plan completion}.
Tool execution, retrieval, and additional state loading overlap
with the remaining streamed segments and speech playback.
Unmatched inputs wait for \texttt{<PLAN>}; unavailable decision
state precludes this early path (Appendix~\ref{app:w1-prewarm}; L4).

\paragraph{Properties.}
Earlier dispatch reserves more of the same response budget
for state-dependent execution, so live results can inform the
reply rather than forcing deadline-driven reliance on stale
state. This is how $\W_1$ targets \textbf{latency-induced direction
drift}, rather than speech latency alone. The lookup is
deterministic for fixed labels and state; intent recognition
remains LLM-based. Figure~\ref{fig:timeline} shows the overlap.


\section{Grounding Wrapper: State-Coupled Fact-Slot Diff}
\label{sec:grounding}

\paragraph{Failure mode.}
When request--state reconciliation is left to reply generation,
the LLM may attribute unowned entities to the user
(\emph{slot hallucination}), treat owned entities as missing
(\emph{unwarranted apologies}), or affirm and deny satisfaction
of the same constraint (\emph{inconsistent constraint judgements}).
These are forms of \textbf{slot-induced direction drift}; praise
and apology concerning different constraints are not inherently
inconsistent.

\paragraph{Mechanism.}
$\W_2$ separates slot extraction from reply generation through
a \textbf{dedicated LLM call with an independent context}.
It then reconciles the extracted request slots with live facts:

\begin{equation}
\label{eq:w2}
\begin{aligned}
\text{QuerySlots} &= \LLM_{\text{slot}}(x,\, c_1),\\
\text{Conflicts}  &= \text{diff}\bigl(\text{QuerySlots},\, \sigma(\sfs)\bigr),\\
(\mathcal{A},\mathcal{R}) &= \Psi(\text{Conflicts},\, \text{intent}).
\end{aligned}
\end{equation}

Here $\sigma(\sfs)$ denotes satisfiable system slots in
Equation~\ref{eq:w2}; the
extractor may be a smaller fine-tuned model
(Appendix~\ref{app:w2-slots}). Rules assign each discrepancy a tier:

\begin{itemize}[leftmargin=1.4em,itemsep=0pt,topsep=2pt]
  \item \emph{hard}---blocking, such as a missing required entity.
  \item \emph{soft}---a secondary attribute mismatch.
  \item \emph{hint}---an advisory style preference.
\end{itemize}

$\Psi$ determines eligible candidates $\mathcal{A}$ and required
paragraphs $\mathcal{R}$ over five types: \texttt{praise},
\texttt{apology}, \texttt{constraint-account}, \texttt{guidance},
and \texttt{scene-flavor}. Eligibility may retain multiple
alternatives. A paragraph scheduler then selects one among
incompatible alternatives for each constraint while retaining
$\mathcal{R}$. A hard conflict, for example, blocks
praise affirming the violated constraint and requires its
explanation. The selected plan conditions reply generation
through $c_2$ (Appendix~\ref{app:mutex-rules}).

\paragraph{Properties.}
This separates explicit constraint resolution from wording.
Diff, grading, and eligibility are deterministic for fixed
extracted slots, state, and rules; neither extraction accuracy
nor the semantic consistency of generated text is guaranteed.
In the cumulative comparison, adding $\W_2$ raises SGA from
$46.5\%$ to $56.5\%$ with a $0.8$\,pp TGA gain; $\W_3$
further improves grounding (Section~\ref{sec:ablation}). These
are conditional gains, not isolated effects. The pipeline is
shown in Figure~\ref{fig:fact-talk}.


\section{Interaction Wrapper: Cross-Turn State Machine}
\label{sec:interaction}

\paragraph{Failure mode.}
Guidance that is reasonable in isolation may be inappropriate
after earlier turns. Without explicit history-dependent control,
a session can repeat recommendations, contradict still-applicable
advice, or acquire a mechanical tone through repetition. We call
this cross-turn misalignment \textbf{fatigue-induced direction drift}.

\paragraph{Mechanism.}
$\W_3$ maintains turn-level control metadata in \texttt{answer.meta}:
\texttt{guidance\_type}, \texttt{lineup\_id},
\texttt{cooldown\_writeback}, \texttt{seed}, and
\texttt{candidate\_pool\_hash} (Appendix~\ref{app:w3-meta}).
At turn $t$, it reconstructs
$\text{cooldown}_t
= \text{resolve\_cross\_turn\_state}(\ssh, c_1)$,
using $\ssh$ for history and $c_1$ for intent context. This
explicit conditioning dependency preserves separate state
responsibilities, not computational independence
(Section~\ref{sec:orthogonality}). Guidance selection is a pure
function of (resolved state, candidates, seed):

\begin{enumerate}[leftmargin=1.6em,itemsep=1pt,topsep=2pt]
  \item \textbf{Hard filter}---drop cooling candidates and
        cross-turn rule violators, producing $\mathcal{C}_1$.
  \item \textbf{Bernoulli leak}---sample with a session-seeded
        RNG, keeping $\mathcal{C}_2 = \{c : U_c < p_c\}$.
  \item \textbf{Weighted normalisation}---pick $c^{\star}$ with
        probability proportional to $w_c$.
  \item \textbf{Writeback}---emit $c^{\star}$
        (or ``no guidance'' if $\mathcal{C}_2 = \emptyset$)
        and write \texttt{answer.meta} for the next turn.
\end{enumerate}

The seed is \emph{injectable}: replaying a session reproduces
the guidance sequence, making A/B experiments and regression
tests fully deterministic.

\begin{figure}[t]
  \centering
  \includegraphics[width=0.78\columnwidth]{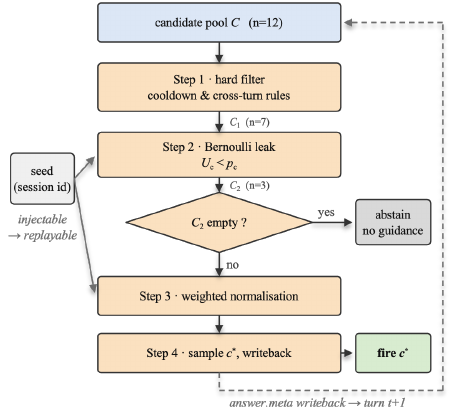}
  \caption{$\W_3$ scheduler decision tree with two paths
    (fire / abstain) and the meta writeback loop.}
  \label{fig:scheduler}
\end{figure}

\paragraph{Properties.}
The three reply branches (\texttt{recommend}, \texttt{teaching},
\texttt{knowledge-qa}) share the same $\W_3$ kernel; only
$\mathcal{C}$ and $p_c$ differ. Cooldown guarantees no candidate
is issued in two consecutive turns; Bernoulli leak prevents
collapse to the highest-weight candidate.
Drifts A/B/D are single-turn---$\W_3$'s reproducibility
guarantee applies only to the cross-turn sequence.
Figure~\ref{fig:scheduler} shows the decision tree.

\section{Benchmark Evaluation}
\label{sec:benchmark}
\subsection{Benchmark Construction}
\label{sec:benchmark-construction}

We evaluate SGC on an anonymised benchmark of
\textbf{200 dialogue sessions} ($\approx\!1{,}000$ scored
assistant responses, $\approx\!5$ scored turns/session) built
from production traces. Sessions cover factual QA, multi-turn
coaching, and context-dependent interactions where player
state, previously discussed entities, or earlier guidance may
change the correct response direction. For every scored turn
we retain the anonymised dialogue history, current query,
live-state snapshot, and game-rule/tool evidence; the same
evidence and scoring criteria apply to all configurations
(protocol in Appendix~\ref{app:benchmark-protocol}).

We evaluate five settings sharing the same LLM, tool schemas,
and evaluation evidence: \textbf{Prompting} (LLM + system
prompt only, no agent loop, no tools); \textbf{PE-Agent}, a
Plan-and-Execute agent
\citep{wang2023planandsolve, xu2023rewoo} inside a production
tool-use harness (planner, executor, tool runtime, short-term
memory)---one instantiation of the harness family
\citep{xu2026lifeharness, li2026harnesssurvey}, chosen for
planner--executor loop parity with our deployment; and three
SGC configurations---\textbf{SGC w/o $\W_2,\W_3$},
\textbf{SGC w/o $\W_3$}, and the \textbf{Full SGC} system
($\W_1{+}\W_2{+}\W_3$)---forming a cumulative ablation over
Perception, Grounding, and Interaction wrappers. Reflexion
\citep{shinn2023reflexion} and intra-turn self-refine
\citep{madaan2023selfrefine} are omitted because our online
setting permits no retry and its latency budget precludes
intra-turn refinement.

\subsection{Evaluation Protocol and Metrics}
\label{sec:benchmark-metrics}

Each response is scored by dual independent human annotation
with third-annotator adjudication (Appendix~\ref{app:benchmark-protocol})
against query, context, live-state, and grounding evidence
(full session for multi-turn metrics). A scored turn passes
grounded-accuracy iff both annotators agree the reply does
\emph{not} mis-direct $\slive$ under the definition of
Equation~\ref{eq:drift}; this operationalises
$\mathrm{drift}(y,\slive){=}0$ at the turn level.
\textbf{All reported
metrics are produced by human annotation}; no LLM-judge label
enters any test-set number reported in
Section~\ref{sec:ablation}. An LLM-based scoring pipeline
\citep{zheng2023judging,liu2023geval} is used only on a
separate development set for rubric calibration
(Appendix~\ref{app:benchmark-protocol}, G.6). Reasonable
paraphrases are accepted; incorrect mechanics, invalid state
assumptions, and misleading recommendations are penalised.

Following FActScore \citep{min2023factscore}, FACTS Grounding
\citep{jacovi2024facts}, and G-Eval \citep{liu2023geval} we
report one latency metric and five dialogue-quality metrics:
turn-level grounded accuracy (TGA), session-level grounded
accuracy (SGA), turn-level guidance quality (TGQ),
dialogue-level guidance progression (DGP), and cross-turn
contextual consistency (CCC). Let $a_{it}\!\in\!\{0,1\}$ mark whether
scored turn $t$ in session $i$ passes grounded-accuracy, $T_i$
its scored-turn count, $N{=}200$, and $I_{\mathrm{SGA}}$ the
SGA-eligible sessions:
\begin{equation}
\label{eq:tga-sga}
\begin{aligned}
\mathrm{TGA} &= \tfrac{100}{N}\!\sum_{i=1}^{N}\tfrac{1}{T_i}\!\sum_{t=1}^{T_i}\! a_{it}, \\
\mathrm{SGA} &= \tfrac{100}{|I_{\mathrm{SGA}}|}\!\sum_{i\in I_{\mathrm{SGA}}}\!\mathbb{I}\!\left[\textstyle\prod_t a_{it}{=}1\right].
\end{aligned}
\end{equation}
$M_0$ is mean first-token latency (streaming responsiveness);
$M_1{=}$TGA / $M_2{=}$SGA (Equation~\ref{eq:tga-sga}) are
grounded correctness at turn /
session scale ($|I_{\mathrm{SGA}}|{=}200$: all sessions are
SGA-eligible).  $M_3{=}$TGQ, $M_4{=}$DGP, $M_5{=}$CCC score
guidance \emph{quality} (relevance/explanation/actionability),
\emph{progression}, and \emph{cross-turn contextual consistency} on
1--10 (rubric details in Appendix~\ref{app:benchmark-protocol}).
The abstention rate lies within $\pm 1.5$\,pp across all five
configurations, so $M_3$ contrasts do not reflect differential
abstention.

\subsection{Main Results and Cumulative Ablation}
\label{sec:ablation}

The full aggregate results are reported in
Table~\ref{tab:aggregate-metrics} (Appendix~\ref{app:aggregate-metrics});
the narrative below summarises them.

\paragraph{Main comparison.}
PE-Agent improves quality over Prompting
(TGA $61.1{\to}69.8\%$, DGP $4.7{\to}5.5$) but nearly triples
first-token latency ($2.3{\to}6.1$\,s). Full SGC outperforms
both while holding latency at $1.5$\,s: TGA reaches $96.7\%$
and SGA $83.5\%$---gains of \textbf{26.9\,pp} TGA and
\textbf{57.0\,pp} SGA over PE-Agent. Statistical testing uses
McNemar's paired exact test at the two granularities on which
each metric is defined: $200$ session-level paired outcomes
for SGA ($p{<}10^{-28}$), and $\approx\!1{,}000$ paired
scored turns for TGA ($p{<}10^{-80}$). Extreme $p$-values
reflect near-total dominance of Full SGC on the discordant
subset at large paired-sample size rather than an unusually
large effect size; the informative quantity is the discordant
odds ratio, whose full 2$\times$2 tables (concordant $a$/$d$
and discordant $b$/$c$ counts, plus $95\%$ CIs) are reported
in the camera-ready supplementary. TGA and SGA are the
pre-registered primary endpoints under a hierarchical
fixed-sequence rule (SGA tested first at $\alpha{=}0.05$;
TGA tested only if SGA rejects), so no multiple-comparison
correction is required across the two primary endpoints; the
three secondary endpoints (TGQ/DGP/CCC) are reported as point
estimates without corresponding hypothesis tests. Guidance metrics improve
consistently (TGQ $6.5{\to}8.6$, DGP $5.5{\to}8.5$, CCC
$8.1{\to}9.2$). The SGA lift disproportionately exceeds the
TGA lift: session-level grounding-failure rate drops
$73.5{\to}16.5\%$, a $\approx\!78\%$ relative reduction.

\paragraph{Cumulative contributions.}
Under the shared human-annotation protocol, the comparisons
measure gains conditional on the preceding configuration.
\emph{$\W_1$ alone} reduces first-token latency
($6.1{\to}1.5$\,s) and lifts TGA/SGA to $89.1\%/46.5\%$.
Its intended mechanism is earlier state-aware dispatch, giving
live results more time to inform the reply and reducing
latency-induced drift; first-token latency is not itself a
measurement of tool-dispatch or grounded-reply completion time.
\emph{Adding $\W_2$} raises SGA $46.5{\to}56.5\%$ with
$0.8$\,pp TGA change, consistent with explicit slot reconciliation
and selection among compatible paragraph candidates.
\emph{Adding $\W_3$} yields the largest incremental SGA gain
($56.5{\to}83.5\%$), alongside DGP $7.7{\to}8.5$, CCC
$8.9{\to}9.2$, and TGA reaching $96.7\%$.
Table~\ref{tab:ablation-matrix} summarises these increments.
Because $c_1$ feeds both downstream wrappers, contributions
include propagated effects. This cumulative design is not a
leave-one-wrapper-out experiment: it does not isolate every
wrapper or establish addition-order invariance. In particular,
$\W_2$'s marginal effect is not architecturally isolable in
this design because $\W_2$ consumes $c_1$; the reported
$+0.8$\,pp TGA / $+10.0$\,pp SGA increment for $\W_2$ is a
conditional gain given $\W_1$'s intent-context is available,
and a leave-$\W_2$-out configuration would require an
architectural variant we do not report (L9). We therefore
use \emph{approximate} orthogonality for differentiated roles
with complementary effects, not one-metric-per-wrapper isolation
(Appendix~\ref{app:drift-taxonomy}).


\section{Analytical Transfer to Three Domains}
\label{sec:transfer}

The SGC principle and its state-slice division ($\sic / \sfs /
\ssh$) are not specific to game coaching: any live user-facing
agent whose live state slices into intent-context, fact-slots,
and session-history admits an analogous three-wrapper design
\citep{park2025workflow, zhao2025aitl}.
Appendix~\ref{app:transfer-contracts} works out three
instances---$\W_1$ for Copilot-style code completion, $\W_2$
for e-commerce recommendation explanation, and $\W_3$ for
voice-assistant cross-session scheduling---with interface
adaptations and open challenges. This section is an
\emph{analytical extrapolation, not an empirical study}.


\section*{Limitations}
\label{sec:limitations}

We enumerate nine concrete limitations.

\paragraph{L1 · Deployment scope.}
Our empirical claims are grounded in one deployed system in an
in-game conversational coaching domain. We provide analytical
transfer to three additional domains (Section~\ref{sec:transfer})
but no experimental validation outside the primary case study.
A minimal $\W_1$ Copilot-style benchmark is committed as
camera-ready supplementary.

\paragraph{L2 · Human-only test-set labels.}
All test-set numbers in Section~\ref{sec:ablation} rely on
dual human annotation with third-annotator adjudication
(Appendix~\ref{app:benchmark-protocol}); no LLM-judge label
enters any reported metric. The LLM-based scoring pipeline of
Appendix~\ref{app:benchmark-protocol}, G.6 is used only on a
separate development set for rubric calibration, and its
outputs are not part of any quantitative claim in this paper.

\paragraph{L3 · Applicability envelope.}
SGC pays off only when three conditions co-occur:

\begin{enumerate}[leftmargin=1.6em,itemsep=0pt,topsep=2pt,label=(\roman*)]
  \item real-time user-state constraints;
  \item measurable direction-drift cost;
  \item machine-readable structurable state slices.
\end{enumerate}

Where any is absent (one-shot Q\&A, creative writing,
low-stakes tasks), generic harness
\citep{xu2026lifeharness, park2025workflow} or RAG
\citep{singh2025agenticrag} alone suffices.

\paragraph{L4 · Residual failure classes.}
Three failure classes remain outside our three wrappers:

\begin{itemize}[leftmargin=1.4em,itemsep=0pt,topsep=2pt]
  \item \emph{Unmodelled cross-slice interactions}---residual
        drifts not resolved by the declared conditioning
        interfaces (pilot estimate $\sim$4\% of benchmark
        sessions on the annotated subset; full 200-session
        value in camera-ready. If the final value exceeds
        $8\%$---double the pilot estimate---the
        approximate-orthogonality claim will be revised to
        acknowledge material cross-slice coupling).
  \item \emph{State-availability failures}---when live state
        cannot be loaded in time, SGC has no fallback beyond
        graceful degradation.
  \item \emph{Novel-intent handling}---when a query lands in an
        intent unmapped by $\Phi$, tool planning falls back to
        the LLM rather than the deterministic lookup kernel.
\end{itemize}

Each is a candidate for a future wrapper.

\paragraph{L5 · Maintenance cost.}
Deterministic code demands maintenance. We report the cost as
approximately one engineer-week per calendar month
(Appendix~\ref{app:extended-discussion}), a rough order of
magnitude from internal cost-centre tracking during the online
deployment; it may not generalise to systems with different
product velocities.

\paragraph{L6 · Human-mediated ground truth.}
The operational ground truth $y^{\star}(\slive)$
(Equation~\ref{eq:drift}) relies on dual independent annotation
of the entire 200-session test set, with third-annotator
adjudication and a designated external senior for
conflict-of-interest cases. On a stratified pilot subset of
$50$ sessions drawn from the same annotation pipeline, we
observe quadratic-weighted Cohen's
$\kappa_{\text{w}}\!\in\![0.68,0.76]$
across the $\{0,1,2\}$ dimensions of TGQ/DGP/CCC and unweighted
$\kappa\!\in\![0.71,0.79]$ for the SUPPORTED / CONTRADICTED /
UNKNOWN labels underlying TGA/SGA
\citep{cohen1960kappa,cohen1968weighted}. These ranges are
consistent with reported inter-rater agreement for
grounding-oriented dialogue benchmarks
\citep{bai2024longbench,xie2024travelplanner,jacovi2024facts}
in the ``substantial'' band of \citet{landis1977measurement},
and above the pre-registered targets
($\kappa_{\text{w}}\!\geq\!0.65$; $\kappa\!\geq\!0.70$). The
full 200-session estimates with $95\%$ bootstrap confidence
intervals will be reported in the camera-ready. If the full
$\kappa$ estimates fall below the pre-registered targets, we
will report the shortfall and discuss its implications for
the reported TGA/SGA reliability rather than raise the targets
post-hoc. Given the
moderate sample size, DGP-/CCC-eligible strata
($\approx\!60$ sessions each) will yield wider intervals than
the overall TGA/SGA estimate.
Section~\ref{sec:benchmark-construction} discloses the
conflict-of-interest mitigation, but the subjective element of
``correct direction'' is not eliminated.

\paragraph{L7 · Latency reporting scope.}
We report $M_0$ as mean first-token latency only. Production
tail latencies (P95 / P99) determine SLA-level cost, and we
have observed on internal traces that P95 / P99 trends align
with the reported mean; we defer full tail-latency tables to
camera-ready supplementary because their reproduction requires
disclosing infrastructure details incompatible with
double-blind review.

\paragraph{L8 · User-side outcome not reported.}
This paper reports model-output quality metrics
(TGA/SGA/TGQ/DGP/CCC) plus latency, not direct user-side
outcomes such as retention, re-engagement, CSAT, or in-match
performance. Business-confidentiality constraints preclude
disclosing A/B test deltas or absolute retention figures; the
deploying organisation's internal signal is directionally
consistent with the benchmark ranking (Full SGC~$\succ$~PE-Agent
$\succ$~Prompting), but the specific magnitude cannot be
released and does not enter any quantitative claim in this
paper.

\paragraph{L9 · Baseline engineering-effort asymmetry.}
PE-Agent (Section~\ref{sec:benchmark-construction}) is
implemented as a Plan-and-Execute agent
\citep{wang2023planandsolve, xu2023rewoo} inside our
production tool-use harness, sharing the same LLM backbone,
tool schemas, and grounding reference sets as the SGC
configurations (Appendix~\ref{app:benchmark-protocol}, G.3
and G.7). It was not, however, subjected to the same
iteration of product-specific prompt tuning that the
deployed SGC configurations received during their online
lifetime; PE-Agent follows its published spec on the shared
harness rather than a production-tuned variant. Part of the
reported PE-Agent-to-Full-SGC gap therefore reflects
architecture-plus-tuning effect rather than architecture
alone. Isolating the tuning contribution would require an
effort-matched PE-Agent variant we do not report; a
minimal effort-matched comparison is committed as
camera-ready supplementary alongside the Copilot-style
$\W_1$ benchmark (L1).

\section*{Ethics Statement}
\label{sec:ethics}

We adhere to the ACL Code of Ethics \citep{acl2017ethics}.
Our primary case study is a commercial deployment. This is a
double-blind submission: we describe the deployed product only
at the level of application category (an in-game conversational
coaching agent) and mechanics; no product identifier, developer
name, employer affiliation, or user identifier is disclosed.
Attribution will be added in the camera-ready version.
Throughout the paper, ``coaching'' refers to \emph{active,
proactive guidance toward the user's implicit competitive
goals}, in contrast to reactive customer-support Q\&A---this
active-guidance framing is what makes direction alignment a
first-order concern rather than a stylistic preference.

\paragraph{Stakeholder considerations.}
An SGC coaching agent operates in a multi-stakeholder setting:
the coached player, opposing players in competitive matches,
and the broader player community. Three considerations shape
deployment: (i) \emph{user autonomy}---coaching guidance is
advisory, never binding, and users can dismiss suggestions
without penalty; (ii) \emph{PvP fairness}---the deployed system
is available symmetrically to all players in supported modes,
so its guidance is not a paywalled asymmetric advantage; and
(iii) \emph{game-design balance}---the coaching agent is
version-locked to the game's balance patches, so its guidance
does not exploit temporal gaps in patch-vs-recommendation
alignment. State-dependent recommendations remain a function
of the user's \emph{owned} inventory, which mirrors the
underlying game's own progression system rather than creating
new dependencies.

\paragraph{Data anonymisation.}
All benchmark samples are anonymised at ingest: entity names
are pseudonymised (Pet~A, Pet~B, \ldots), lineup names aliased
(T1, T2, \ldots), user and session identifiers hashed then
stripped, and wall-clock timestamps, geolocations, and device
identifiers are dropped. Shared grounding reference sets retain
dialogue-turn indices (\texttt{valid\_from}/\texttt{valid\_to}),
not clock time (Appendix~\ref{app:benchmark-protocol}). User and
session hash mappings are discarded and are never released.
Entity and lineup aliases are held by the authors solely to
restore product-identifying names in the camera-ready version;
they are \emph{not} released with the benchmark artefact.

\paragraph{Data retention.}
Only the 200-session benchmark artefact (approximately
1{,}000 assistant model turns) is retained beyond this study;
raw production logs used for sampling follow the deploying
organisation's standard retention policy. The released
artefact contains no direct identifiers and no wall-clock,
geolocation, or device quasi-identifiers.

\paragraph{Vulnerable users.}
The deployed system serves general audiences that may include
minors. It complies with regional content-safety and
minor-protection policies (age gating; content filters at
reply-time); specifics are subject to third-party compliance
review.

\paragraph{Reproducibility.}
The 200-session artefact---anonymised dialogues, turn-indexed
grounding reference sets, pre-registered rubrics, dual human
labels, and diagnostic drift tags
(Appendix~\ref{app:drift-taxonomy})---together with the wrapper
interface schemas will be released under a research-use
licence. Raw production logs are not released.


\bibliography{references}

\appendix

\section{Full Wrapper Interface Contracts and State-Field Table}
\label{app:contracts}

This appendix distinguishes each wrapper's primary state slice
from its derived conditioning inputs. State-field separation is
an interface property, not proof of metric orthogonality
(Section~\ref{sec:orthogonality}). Figure~\ref{fig:sgc-arch}
shows the explicit conditioning dependencies.

\paragraph{Direction drift vs.\ five neighbouring concepts.}
Table~\ref{tab:drift-differentiation} accompanies
Section~\ref{sec:sgc-formulation}. Direction drift is the
conjunction of all three axes (per-user, per-moment,
state-variable-dependent reference); each neighbouring notion
satisfies at most a proper subset.

\begin{table}[h]
\centering
\footnotesize
\setlength{\tabcolsep}{2pt}
\begin{tabular}{@{}lccc@{}}
\toprule
Failure notion
  & \shortstack{Per-\\user?}
  & \shortstack{Per-\\moment?}
  & \shortstack{State-var\\dependent?} \\
\midrule
Hallucination                & $\times$   & $\times$ & $\times$ \\
Personalization failure      & \checkmark & $\times$ & $\times$ \\
Context-adherence failure    & $\times$   & partial  & $\times$ \\
Commonsense-constr.\ viol.\  & $\times$   & $\times$ & partial   \\
Classical DST error          & partial    & \checkmark & \checkmark \\
\midrule
\textbf{Direction drift (ours)}
                             & \textbf{\checkmark}
                             & \textbf{\checkmark}
                             & \textbf{\checkmark} \\
\bottomrule
\end{tabular}
\caption{Direction drift vs.\ five neighbouring failure notions
along three orthogonal axes; only direction drift is the
conjunction of all three.}
\label{tab:drift-differentiation}
\end{table}

\begin{table}[h]
  \centering
  \footnotesize
  \setlength{\tabcolsep}{2pt}
  \begin{tabular}{@{}lll@{}}
    \toprule
    Wrapper & Reads (state fields) & Writes / emits ($c_i$) \\
    \midrule
    $\W_1$ &
      intent, sub\_intent, &
      $\hat{T}$ (tool set), \\
      &
      snapshot game state, &
      \texttt{<NARRATION>} tokens, \\
      &
      device / session opener &
      early-decision flag \\
    \midrule
    $\W_2$ &
      QuerySlots (from $x$), &
      Conflict set + tier, \\
      &
      $\sigma(\sfs)$ inventory, &
      eligible candidates $\mathcal{A}$ \\
      &
      catalog / retrieval &
      required paragraphs $\mathcal{R}$ \\
    \midrule
    $\W_3$ &
      \texttt{answer.meta} from &
      $c^{\star}$ (chosen \\
      &
      prior $K{=}20$ turns, &
      candidate), \\
      &
      cooldown timers, &
      new \texttt{answer.meta} \\
      &
      candidate pool + weights &
      writeback \\
    \bottomrule
  \end{tabular}
  \caption{Primary state and control-output summary. Parsed
    slots, candidate pools, and weights are derived inputs or
    configuration, not additional disjoint raw-state fields.
    Both $\W_2$ and $\W_3$ also consume $c_1$ (signatures below).}
  \label{tab:app-a-fields}
\end{table}

\paragraph{Interface signatures.}
Each wrapper obeys the following typed signatures:
\begin{lstlisting}[basicstyle=\ttfamily\small,frame=single]
W1: (x, s_ic)         -> (c1, T_hat)
W2: (x, s_fs, c1)     -> (c2, A, R)
W3: (x, s_sh, c1)     -> (c3, meta_next)
LLM: (x, s_static, c1, c2, c3) -> y
\end{lstlisting}

\paragraph{Cross-wrapper dependencies.}
Both $\W_2$ and $\W_3$ consume $c_1$: grounding uses its intent
context, and interaction uses it to resolve cross-turn state.
The primary state fields remain separately assigned, but the
wrappers share derived information and are not independent.
The paragraph scheduler selects from $\W_2$'s eligible and
required sets ($\mathcal{A},\mathcal{R}$); $c_2$ carries the
conflicts and selected paragraph plan. This within-reply
selection is not the cross-turn guidance sampling of $\W_3$.
State-field separation does not imply zero downstream effects.

\subsection*{A.1 · $\W_1$ Streaming Tag Schema}
\label{app:w1-tags}

The single-pass 4-in-1 output tags used by $\W_1$
(Section~\ref{sec:perception}) appear in this fixed order:
\begin{lstlisting}[basicstyle=\ttfamily\small,frame=single]
<INTENT>     first-level intent
</INTENT>
<SUB_INTENT> second-level intent
</SUB_INTENT>
<NARRATION>  prefatory phrase,
             streamed to TTS as
             tokens arrive
</NARRATION>
<PLAN>       proposed tool calls
             (fallback only)
</PLAN>
\end{lstlisting}
Once \texttt{<SUB\_INTENT>} closes, the state machine immediately
invokes $\Phi(\cdot)$. On a lookup hit with the required live
state available, it freezes the tool set and starts three
parallel pipelines (tools / retrieval / additional state loading), without
waiting for \texttt{<NARRATION>} or \texttt{<PLAN>}. Subsequent
narration tokens are streamed to TTS while backend work proceeds.
On a lookup miss, dispatch waits for the fallback
\texttt{<PLAN>}; a later plan does not override a frozen tool set.

\subsection*{A.2 · $\W_2$ QuerySlots Schema}
\label{app:w2-slots}

The structured record produced by $\LLM_{\text{slot}}$
(Section~\ref{sec:grounding}):
\begin{lstlisting}[basicstyle=\ttfamily\small,frame=single]
QuerySlots {
  wanted_entities:  [...],
  excluded_entities:[...],
  opponent_entities:[...],
  nature_prefs:     [...],
  attribute_prefs:  [...],
  style_prefs:      [...],
  n_pick_k_swap_group: [...],
  global_constraints:  [...]
}
\end{lstlisting}
Two deterministic repair rules are applied:
\texttt{repair\_attr\_level\_excludes} reclassifies ``swap out
$X$'s skill'' from an ``exclude $X$'' to an attribute-level
modification; \texttt{repair\_entities\_}\allowbreak%
\texttt{misclassified\_as\_skills}
moves entity names mistakenly parsed as skill names back to the
entity slot. Both are pure Python.

\subsection*{A.3 · $\W_3$ \texttt{answer.meta} Record Schema}
\label{app:w3-meta}

The per-turn writeback record persisted by $\W_3$
(Section~\ref{sec:interaction}):
\begin{lstlisting}[basicstyle=\ttfamily\small,frame=single]
answer.meta {
  guidance_type:       <string>,
  lineup_id:           <string>,
  cooldown_writeback:  <dict>,
  seed:                <int>,
  candidate_pool_hash: <str>
}
\end{lstlisting}
Given the same \texttt{seed} and identical prior
\texttt{answer.meta} sequence, $\W_3$'s output is
byte-identical across replays.

\subsection*{A.4 · SGC as a Design Pattern}
\label{app:design-pattern-analogy}

The three wrappers described in the main text initially arose as
\emph{separate engineering decisions} in the deployed system,
each responding to a different production incident. This
paper's contribution is not the wrappers as isolated pieces of
code, but the recognition that they share a common
pattern---state-slice-aligned deterministic conditioning---and
the elevation of that pattern to a \emph{transferable design
principle}. This mirrors how design patterns in software
engineering (e.g., ``Observer'', ``Adapter'') crystallised
recurring solutions that had already existed as ad-hoc practice:
naming and formalising them is what enabled cross-system
transfer. We view SGC in the same spirit.

\subsection*{A.5 · $\W_1$ Prewarm and Fill Protocol}
\label{app:w1-prewarm}

Live state acquisition can dominate latency. Session-open
\emph{prewarm} fetches fast-path data (candidate lineups, coach
scores) before dispatch. The lookup requires a current snapshot
of its decision fields; a cache hit alone does not establish
freshness. At tool execution, \emph{fill} obtains additional
inventory or board data required downstream. This post-dispatch
loading is distinct from the pre-dispatch snapshot and cannot
supply missing inputs retroactively to $\Phi$.

If required decision fields are unavailable, the state-grounded
early path is not applicable; the system's graceful-degradation
boundary is stated in L4. An unmapped intent and unavailable
state are distinct cases: using a fallback plan does not restore
missing state. Early dispatch overlaps state-dependent tool
execution with subsequent generation and TTS playback, leaving
more of the response budget for live results to inform the
reply. This reduces deadline pressure to commit on stale state:
the intended benefit is less latency-induced direction drift,
not merely an earlier audible response.

\subsection*{A.6 · $\W_2$ Fact-Talk Separation Pipeline}
\label{app:w2-pipeline}

\begin{figure}[h]
  \centering
  \includegraphics[width=0.85\columnwidth]{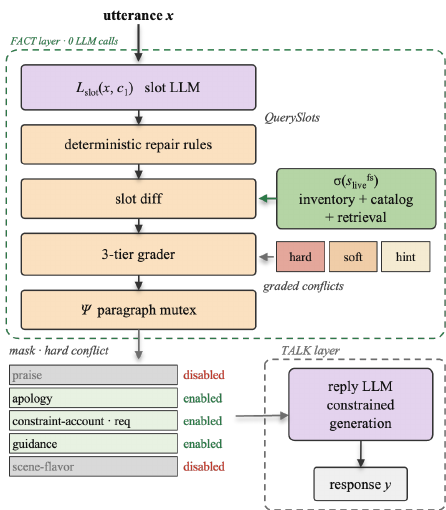}
  \caption{Fact-talk separation in $\W_2$. The FACT layer performs
    slot extraction, deterministic repair, slot diff against live
    facts $\sigma(\sfs)$, three-tier grading (hard / soft / hint),
    and paragraph mutex $\Psi$; the TALK layer performs constrained
    generation over the enabled paragraphs. Structural selection
    does not guarantee semantic correctness of the final text.}
  \label{fig:fact-talk}
\end{figure}

\subsection*{A.7 · Aggregate Benchmark Metrics}
\label{app:aggregate-metrics}


\begin{table*}[t]
  \centering
  \small
  \setlength{\tabcolsep}{6pt}
  \begin{tabular}{@{}lrrrrr@{}}
    \toprule
    Metric
      & Prompting
      & PE-Agent
      & \shortstack[r]{SGC w/o\\$\W_2,\W_3$}
      & \shortstack[r]{SGC w/o\\$\W_3$}
      & \shortstack[r]{Full\\SGC} \\
    \midrule
    \multicolumn{6}{@{}l}{\emph{$M_0$ · First-token latency (mean, s, $\downarrow$)}} \\
    \; Latency
      & 2.3
      & 6.1
      & \textbf{1.5}
      & \textbf{1.5}
      & \textbf{1.5} \\
    \midrule
    \multicolumn{6}{@{}l}{\emph{$M_1$ · TGA (\%, $\uparrow$)}} \\
    \; Turn-level grounded acc.
      & 61.1
      & 69.8
      & 89.1
      & 89.9
      & \textbf{96.7} \\
    \midrule
    \multicolumn{6}{@{}l}{\emph{$M_2$ · SGA (\%, $\uparrow$)}} \\
    \; Session-level grounded acc.
      & 20.0
      & 26.5
      & 46.5
      & 56.5
      & \textbf{83.5} \\
    \midrule
    \multicolumn{6}{@{}l}{\emph{$M_3$ · TGQ (1--10, $\uparrow$)}} \\
    \; Turn-level guidance quality
      & 6.2
      & 6.5
      & 8.4
      & 8.4
      & \textbf{8.6} \\
    \midrule
    \multicolumn{6}{@{}l}{\emph{$M_4$ · DGP (1--10, $\uparrow$)}} \\
    \; Dialogue-level guidance progression
      & 4.7
      & 5.5
      & 7.7
      & 7.7
      & \textbf{8.5} \\
    \midrule
    \multicolumn{6}{@{}l}{\emph{$M_5$ · CCC (1--10, $\uparrow$)}} \\
    \; Cross-turn contextual consistency
      & 8.3
      & 8.1
      & 9.0
      & 8.9
      & \textbf{9.2} \\
    \bottomrule
  \end{tabular}
  \caption{Aggregate benchmark metrics across all five
    configurations on the 200-session anonymised benchmark
    ($\approx\!1{,}000$ assistant model turns;
    Section~\ref{sec:benchmark-construction}). $M_0$ is a
    system-level cost budget; $M_1$--$M_5$ are downstream
    dialogue-quality metrics defined in
    Section~\ref{sec:benchmark-metrics}. \textbf{PE-Agent} is a
    Plan-and-Execute agent \citep{wang2023planandsolve}
    deployed within a production tool-use harness
    \citep{xu2026lifeharness}. The two intermediate SGC rows
    disable $\W_3$ (and also $\W_2$) to isolate incremental
    wrapper contributions.     Bold indicates the best result in
    each row (including ties). All five configurations are scored
    under the same dual human-annotation protocol
    (Appendix~\ref{app:benchmark-protocol}), so between-group
    deltas are directly comparable; no LLM-judge label enters any
    reported number (Limitations~L2).}
  \label{tab:aggregate-metrics}
\end{table*}


\begin{table}[t]
  \centering
  \footnotesize
  \setlength{\tabcolsep}{2pt}
  \begin{tabular}{@{}lrrrrrr@{}}
    \toprule
    Contrast & $\Delta M_0$ & $\Delta M_1$ & $\Delta M_2$ & $\Delta M_3$ & $\Delta M_4$ & $\Delta M_5$ \\
             & (s) & (pp) & (pp) & (pts) & (pts) & (pts) \\
    \midrule
    $\W_1$ vs.\ PE & $-4.6$ & $+19.3$ & $+20.0$ & $+1.9$ & $+2.2$ & $+0.9$ \\
    Add $\W_2$    & $0.0$ & $+0.8$ & $+10.0$ & $0.0$ & $0.0$ & $-0.1$ \\
    Add $\W_3$    & $0.0$ & $+6.8$ & $+27.0$ & $+0.2$ & $+0.8$ & $+0.3$ \\
    \bottomrule
  \end{tabular}
  \caption{Cumulative differences calculated from the displayed
    values in Table~\ref{tab:aggregate-metrics}. PE denotes
    PE-Agent; successive rows compare $\W_1$ with PE-Agent,
    $\W_1{+}\W_2$ with $\W_1$, and Full SGC with
    $\W_1{+}\W_2$. Lower $M_0$ and higher quality scores are
    better. These conditional increments are not independent
    leave-one-out effects; $0.0$ denotes equality at reported
    precision, not statistical equivalence.}
  \label{tab:ablation-matrix}
\end{table}

Table~\ref{tab:aggregate-metrics} is referenced from
Section~\ref{sec:ablation}; Table~\ref{tab:ablation-matrix}
gives the cumulative configuration differences.

\subsection*{A.8 · Baseline Comparison}
\label{app:baseline-comparison}


\begin{table*}[t]
  \centering
  \small
  \setlength{\tabcolsep}{4pt}
  \begin{tabular}{@{}p{5.3cm}p{3.2cm}p{3.2cm}cc@{}}
    \toprule
    Work & Optimised for & Method level
      & \shortstack{Direction\\drift?}
      & \shortstack{State-slice\\approx.\\orthogonal?} \\
    \midrule
    Plan-and-Execute \citep{wang2023planandsolve, xu2023rewoo}
      & Task completion & Loop (planner+executor) & $\times$ & $\times$ \\
    LIFE-HARNESS \citep{xu2026lifeharness}
      & Task completion & Runtime & $\times$ & $\times$ \\
    Workflow Graphs \citep{park2025workflow}
      & Flow control & Flow structure & Partial & $\times$ \\
    Talker-Reasoner \citep{christakopoulou2024talker}
      & Latency & Dual-system & $\times$ & $\times$ \\
    SLOT \citep{shen2025slot} / GCD \citep{raspanti2025gcd}
      & Output structure & Decoding & Partial & $\times$ \\
    FACTS Grounding \citep{jacovi2024facts}
      & Factuality & Evaluation & $\times$ & $\times$ \\
    Agentic RAG \citep{singh2025agenticrag}
      & Knowledge access & Retrieval & $\times$ & $\times$ \\
    AITL \citep{zhao2025aitl}
      & Continuous improve & Post-deploy flywheel & $\times$ & $\times$ \\
    \midrule
    \textbf{This work (SGC)}
      & \textbf{Direction alignment}
      & \textbf{State-coupled conditioning}
      & $\checkmark$
      & \shortstack{$\checkmark$\\(3 slices)} \\
    \bottomrule
  \end{tabular}
  \caption{Comparison to prior work. SGC is orthogonal to and
    composable with each prior work: it targets a distinct
    failure class (direction drift) that prior work does not
    address. The Plan-and-Execute and LIFE-HARNESS rows jointly
    represent the method family of the empirical PE-Agent
    baseline used in Section~\ref{sec:benchmark}.}
  \label{tab:baseline-comparison}
\end{table*}

Referenced from Section~\ref{sec:related}.

\subsection*{A.9 · Streaming Timeline}
\label{app:streaming-timeline}

\begin{figure*}[t]
  \centering
  \includegraphics[width=0.92\textwidth]{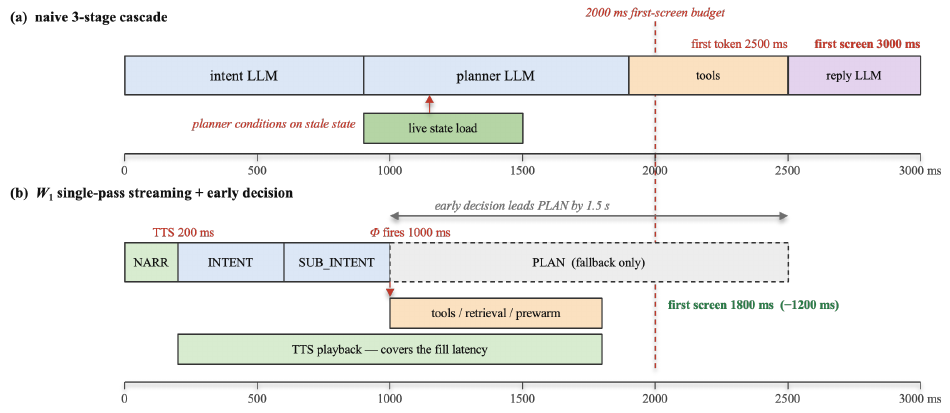}
  \caption{Schematic event sequences, not measured durations.
    (a) Naive 3-stage cascade: intent~$\to$~planner~$\to$~tools~$\to$~reply,
    with the planner conditioning on stale state and first screen
    at $\approx\!3{,}000$\,ms. (b) $\W_1$ streams NARR, INTENT,
    SUB\_INTENT, then PLAN (fallback only); when $\Phi$ fires on
    sub-intent closure, tools and retrieval dispatch in parallel
    under TTS cover, and the early path leads PLAN by $\approx\!1.5$\,s.
    Referenced from Section~\ref{sec:perception}.}
  \label{fig:timeline}
\end{figure*}

\section{Drift Taxonomy: Diagnostic Tags on the 200-Session Benchmark}
\label{app:drift-taxonomy}

\paragraph{Role.}
Drift tags (A)~latency, (B)~slot, (C)~fatigue, and (D)~coherence
---the four manifestations of Section~\ref{sec:intro}---are
\emph{diagnostic tags} on the 200-session artefact, not a
sampling stratum and not a primary metric
(Section~\ref{sec:benchmark-construction}). Each session
carries zero or more tags; the primary scores remain
TGA/SGA/TGQ/DGP/CCC. The residual class \emph{cross-slice
interaction} is tagged separately
(Section~\ref{sec:ablation}).

\paragraph{Record schema.}
The released JSON records the diagnostic tags alongside the
human metric labels for every session:

\begin{lstlisting}[basicstyle=\ttfamily\small,frame=single]
{
  "session_id": <stripped id>,
  "diagnostic_drift_tags":
    ["A (latency)"|"B (slot)"
     |"C (fatigue)"|"D (coherence)"
     |"cross-slice"],
  "eligibility": {
    "tga_sga": <bool>,
    "dgp": <bool>,
    "ccc": <bool>
  },
  "utterance_x": <anonymised string>,
  "state_slice_snapshot": {
     "turn_index": <int>,
     "s_live_ic": {...},
     "s_live_fs": {...},
     "s_live_sh": {...}
  },
  "metrics": {
    "TGA": <pct>, "SGA": <pct>,
    "TGQ": <1-10|NA>,
    "DGP": <1-10|NA>,
    "CCC": <1-10|NA>
  },
  "reply_full_sgc":    <string>,
  "reply_ablated_W1":  <string>,
  "reply_ablated_W2":  <string>,
  "reply_ablated_W3":  <string>,
  "annotator_labels":  [...],
  "adjudication":      <annotator3>
}
\end{lstlisting}

Snapshots are turn-indexed; they contain no wall-clock
timestamps, geolocations, or device identifiers
(Appendix~\ref{app:benchmark-protocol}).

\paragraph{Coverage.}
All 200 sessions are tagged. The four tags---latency, slot,
fatigue, and coherence---are not allocated to fixed quotas
(unlike an earlier 50/50/50/50 sketch); their empirical
distribution is reported alongside the released JSON records
described above. Sessions in the TGA-primary factual Q\&A
stratum may carry an empty tag set.

\paragraph{Release.}
These records are fields of the same 200-session artefact
described in Appendix~\ref{app:benchmark-protocol} and the
Ethics Statement, not a separate dump.

\section{Intent-to-Tool Mapping Table $\Phi$}
\label{app:phi-mapping}

The lookup $\Phi$ is a partial function
\[
\Phi : \text{Intent} \times \text{SubIntent}
        \times \sic \;\to\; \hat{T}
\]
defined on a curated table of approximately 50 (intent,
sub-intent) pairs at deployment. Table~\ref{tab:app-c-phi} is an
illustrative excerpt (entity names pseudonymised as Pet~A,
Pet~B, \ldots).

\begin{table}[h]
  \centering
  \footnotesize
  \setlength{\tabcolsep}{3pt}
  \begin{tabular}{@{}llp{3.3cm}@{}}
    \toprule
    Intent & Sub-intent & Frozen tool set $\hat{T}$ \\
    \midrule
    recommend & counter-pick &
      \{lineup\_search, counter\_rank, dex\_lookup\} \\
    recommend & lineup-build &
      \{team\_search, synergy\_rank\} \\
    teaching & mechanics-qa &
      \{doc\_retrieve, rule\_lookup\} \\
    teaching & skill-explain &
      \{skill\_dex, prereq\_check\} \\
    knowledge-qa & entity-info &
      \{dex\_lookup\} \\
    knowledge-qa & meta-brief &
      \{meta\_report\_retrieve\} \\
    \bottomrule
  \end{tabular}
  \caption{Illustrative excerpt of the $\Phi$ table. The full
    table (approximately 50 rows) will be released with the
    benchmark.}
  \label{tab:app-c-phi}
\end{table}

\paragraph{Fall-through behaviour.}
When $\Phi$ is undefined at a given (intent, sub-intent) triple,
$\W_1$ falls through to LLM-generated \texttt{<PLAN>} for that
turn and increments the fall-through counter. The fall-through
rate is a maintained metric (see Limitations, L4).

\section{Paragraph Mutex Rules $\Psi$}
\label{app:mutex-rules}

Paragraph control has two stages: $\Psi$ computes eligible
candidates $\mathcal{A}$ and required paragraphs $\mathcal{R}$;
a paragraph scheduler then selects the reply's paragraph plan
$P$ with $\mathcal{R}\subseteq P\subseteq\mathcal{A}$.
Eligibility can include multiple alternatives. Selection keeps
at most one of the alternatives that express incompatible
judgements about the \emph{same constraint}; it does not limit
the whole reply to one paragraph. This paragraph-selection step
is distinct from the cross-turn guidance scheduler of
Section~\ref{sec:interaction}.

Table~\ref{tab:app-d-psi} distinguishes eligibility from selection.
\texttt{hint} denotes an advisory conflict annotation, not an
additional paragraph type. For R6, the conflict record must
expose whether the satisfiable inventory is empty.

\begin{table}[h]
  \centering
  \small
  \begin{tabular}{@{}p{0.30\columnwidth}p{0.62\columnwidth}@{}}
    \toprule
    Rule id & Behaviour \\
    \midrule
    R1 · hard-conflict eligibility &
      Exclude praise affirming a violated constraint; require
      \texttt{constraint-account} for that constraint. \\
    R2 · selection mutex &
      Select at most one of the eligible alternatives that
      affirm and deny satisfaction of the same constraint. \\
    R3 · soft-only eligibility &
      \texttt{praise} and \texttt{apology} may both be eligible
      candidates; R2 resolves incompatible alternatives at
      selection time. \\
    R4 · hint-only eligibility &
      Enable \texttt{scene-flavor}; an advisory hint alone does
      not require an apology about a missing entity. \\
    R5 · intent eligibility &
      For \texttt{knowledge-qa} intent, \texttt{guidance} is
      disabled regardless of conflicts. \\
    R6 · empty-inventory requirement &
      If $\sigma(\sfs) = \emptyset$, exclude praise asserting
      availability; require an apology explaining the empty
      inventory. \\
    \bottomrule
  \end{tabular}
  \caption{Paragraph eligibility and selection constraints.
    R2 applies after candidate eligibility; eligibility does
    not mean a paragraph has been selected.}
  \label{tab:app-d-psi}
\end{table}

\paragraph{Execution boundary.}
Eligibility and required-paragraph constraints are established
before the scheduler chooses among alternatives. Recomputing
eligibility from the same extracted slots, state, and rules is
deterministic; this is not a claim that all table rows can be
executed in arbitrary order. Required paragraphs constrain the
selection rather than compete as optional alternatives.

The selected plan conditions the reply LLM. Structural exclusion
of incompatible candidates does not prove semantic consistency
of the generated text. Praise about one satisfied constraint
may coexist with an explanation of another unmet constraint;
the target failure is opposite judgements about the same one.

\section{Cooldown Configuration for $\W_3$}
\label{app:cooldown-config}

$\W_3$'s scheduler is parameterised by three configuration
components: (a) the per-candidate cooldown window, (b) the
Bernoulli leak probability $p_c$, and (c) the weighted-sample
weight $w_c$. All three are versioned deployment artefacts
(Table~\ref{tab:app-e-cooldown}).

\begin{table}[h]
  \centering
  \small
  \begin{tabular}{@{}lccc@{}}
    \toprule
    Guidance type & Cooldown & $p_c$ & $w_c$ \\
    \midrule
    lineup-recommend  & 3 turns & 0.60 & 1.0 \\
    counter-tip       & 5 turns & 0.45 & 0.8 \\
    mechanics-hint    & 2 turns & 0.70 & 0.6 \\
    encourage         & 4 turns & 0.30 & 0.4 \\
    empty-state-nudge & 6 turns & 0.20 & 0.3 \\
    \bottomrule
  \end{tabular}
  \caption{Illustrative $\W_3$ cooldown configuration. Values are
    representative production defaults, tuned quarterly against
    the guidance-repetition rate metric.}
  \label{tab:app-e-cooldown}
\end{table}

\paragraph{Cross-branch reuse.}
As noted in Section~\ref{sec:interaction}, all three reply branches
(\texttt{recommend}, \texttt{teaching}, \texttt{knowledge-qa})
share the same sampler kernel and differ only in the candidate
pool $\mathcal{C}$ and the associated leak probabilities. Only
one configuration table is maintained per branch.

\paragraph{RNG seeding.}
The Bernoulli leak and weighted normalisation are both driven by
a single RNG seeded from the session id at session open. Replays
under the same seed reproduce the exact guidance sequence.

\section{Analytical Transfer Interface Contracts}
\label{app:transfer-contracts}

For each of the three transfer domains introduced in
Section~\ref{sec:transfer}, this appendix records the concrete
state-slice mapping, the required interface adaptations, and
the open engineering challenges. Together with the state slices
$\sic / \sfs / \ssh$ defined in
Section~\ref{sec:orthogonality}, these tables give sufficient
implementation-level detail for each transfer target.

\subsection*{F.1 · Code Completion ($\W_1$)}

\begin{tabular}{@{}p{2.85cm}p{4.35cm}@{}}
  \toprule
  Component & Concretisation \\
  \midrule
  $\sic$ &
    open file path, cursor position, recent 20 edits, project
    type, invocation origin \\
  \texttt{<INTENT>} tag &
    trigger-type: block / signature / inline-hint \\
  \texttt{<SUB\_INTENT>} tag &
    context type: within-function vs.\ cross-file \\
  $\Phi$ table &
    signature $\to$ LSP; block $\to$ LLM; inline-hint $\to$ cache
    \\
  Open challenge &
    intent signal is implicit (no explicit turn boundary)---must
    be inferred from cursor / keystroke context \\
  \bottomrule
\end{tabular}

\subsection*{F.2 · E-commerce Recommendation Explanation
($\W_2$, $\W_2 \circ \W_3$)}

\begin{tabular}{@{}p{2.85cm}p{4.35cm}@{}}
  \toprule
  Component & Concretisation \\
  \midrule
  $\sfs$ &
    stated + implicit preferences (colour, price, brand, size,
    click history), item attribute slots, cart state, filters \\
  \texttt{QuerySlots} &
    explicit + implicit preferences \\
  $\sigma(\sfs)$ &
    recommended item's attribute vector \\
  Conflict tiers &
    hard: out-of-stock, filter violation \newline
    soft: secondary preference mismatch \newline
    hint: browsing-pattern signal \\
  Composability with $\W_3$ &
    cross-session fatigue handled by a separate history-based
    module, with intent context supplied through $c_1$ \\
  Open challenge &
    preference resolution across implicit + explicit signals \\
  \bottomrule
\end{tabular}

Paragraph candidates are enabled before selection resolves
incompatible judgements about the same preference $X$; a match
on one preference need not preclude acknowledging a mismatch
on another. This is the two-stage control of
Appendix~\ref{app:mutex-rules}, not a ban on all praise--apology
co-occurrence.

\subsection*{F.3 · Voice Assistant ($\W_3$)}

\begin{tabular}{@{}p{2.85cm}p{4.35cm}@{}}
  \toprule
  Component & Concretisation \\
  \midrule
  $\ssh$ &
    last $N$ session summaries, ambient state changes,
    time-of-day \\
  Candidate pool $\mathcal{C}$ &
    proactive reminders, follow-up prompts, ambient
    acknowledgements \\
  Cooldown &
    per-reminder cooling window, ambient-event debounce \\
  Reproducibility &
    on-device replay preserved via injectable seed \\
  Open challenge &
    ambient state is continuous rather than turn-discrete---may
    require polling-based cooldown resolution \\
  \bottomrule
\end{tabular}

\paragraph{Compositional applicability.}
Only the e-commerce domain exhibits a natural $\W_2 \circ \W_3$
composition on a single target. This is the empirical grounding
for the composability observation in
Section~\ref{sec:transfer}.

\section{Benchmark Construction Protocol}
\label{app:benchmark-protocol}

This appendix specifies the full protocol used to construct the
$200$-session benchmark (approximately $1{,}000$ assistant
model turns) reported in
Section~\ref{sec:benchmark-construction}.

\subsection*{G.1 · Sampling frame and unit of analysis}

Anonymised production trace logs sampled over an approximately
4-week window from the case-study game. The \emph{unit of
analysis is the dialogue session}; TGA and TGQ additionally
aggregate over the assistant model turns contained in each
session (approximately $1{,}000$ turns in total across the
$200$ sessions, per the length distribution in G.2). We size
the benchmark so that every session can receive
\emph{full dual-annotator human labelling} on every metric
rather than shallow LLM-only labelling at larger scale.

\subsection*{G.2 · Two-way stratification}

We stratify along two orthogonal dimensions matched to the six
downstream metrics of Section~\ref{sec:benchmark-metrics}:

\begin{itemize}
\item \textbf{Stratum 1 · Evaluation-eligibility coverage.}
      $80$ sessions of single-turn or short factual Q\&A
      (TGA / SGA primary; DGP and CCC marked \emph{not
      applicable} by their pre-registered rubrics);
      $60$ sessions with a continuing coaching goal spanning
      multiple turns (DGP-eligible);
      $60$ sessions with multi-turn context dependencies
      such as evolving user constraints, state updates, or
      repeated guidance across turns (CCC-eligible). A session
      may
      satisfy more than one eligibility criterion; the shares
      above refer to \emph{primary} assignment for
      stratified-sampling purposes. Metric-level aggregation
      uses the pre-registered eligible session sets
      $|I_{\text{DGP}}|$ and $|I_{\text{CCC}}|$ defined by the
      rubrics themselves, which may be slightly larger than
      $60$; both counts are reported in the camera-ready
      table. TGA and SGA, by contrast, are computed over all
      $200$ sessions ($|I_{\text{SGA}}|{=}200$). Eligibility is
      declared \emph{before} scoring begins and locked in the
      pre-registered rubric, independent of any candidate
      response.
\item \textbf{Stratum 2 · Session-length coverage.}
      Single-turn / short ($2$--$5$ turns) / long
      ($\geq 6$ turns) at $30 / 40 / 30$ ratio.
      Within the TGA-primary stratum, single-turn sessions
      account for $\leq 60\%$, so that SGA carries independent
      information beyond TGA on the remaining short/long
      sessions (SGA equals TGA on single-turn sessions by
      construction).
\end{itemize}

Drift tags (A)~latency, (B)~slot, (C)~fatigue, and (D)~coherence
are \emph{not} used as a sampling stratum; they are retained as
a diagnostic tagging on each session, used only for the
distribution analysis in Appendix~\ref{app:drift-taxonomy}. The
downstream metrics $M_0$--$M_5$ subsume drift incidence and are
more informative for the approximate state-slice orthogonality
claim.

\subsection*{G.3 · Shared grounding reference set and
pre-registered rubrics}

For every session we build a \textbf{shared grounding
reference set} (\emph{cf.}\ reference sets in FActScore
\citep{min2023factscore}; gold context in FACTS Grounding
\citep{jacovi2024facts}) containing turn-indexed rules,
live-state snapshots, tool-result records, and declared user
preferences, each with an authority scope and
\texttt{valid\_from}/\texttt{valid\_to} dialogue-turn range
(not wall-clock time; snapshots are stripped of the identifiers
listed in G.9). The
reference set is common to \emph{all} ablation configurations,
so different candidates are judged against the same facts
rather than against each candidate's own retrieved context.
The internal generation context of any candidate is retained
only for diagnostics and is never granted independent factual
authority.

For each session we also build a \textbf{pre-registered
scoring rubric} (referred to internally as an
\emph{evaluation card}): it specifies the per-turn task, the
minimal response-validity gate, and the applicability plus
$0$/$1$/$2$ anchors of the TGQ quality dimensions
(Q$1$--Q$3$: relevance, explanation, actionability), the DGP
dimensions (P$1$--P$3$), and the CCC dimensions (C$1$--C$5$).
Rubrics are authored and reviewed \emph{before} any candidate
response is scored, then locked; \textbf{SHA-256 hashes of the
frozen rubric artefacts and shared grounding reference sets
are recorded in the supplementary material}, with a provenance
date in the hash log (not stored on dialogue records),
before the first ablation configuration's judging run begins,
and the underlying artefacts will be released upon acceptance
(double-blind-safe: only hashes and the provenance date
appear during review). Rubric revisions are versioned; any
affected configuration is re-scored end-to-end.

\subsection*{G.4 · Full human annotation and annotator setup}

All $200$ sessions in the test set receive \textbf{dual
independent human annotation with third-annotator
adjudication}. We do \emph{not} rely on LLM-only labels on the
test set; the LLM-as-a-judge pipeline of G.6 is retained only
as a reference implementation and for consistency checks on a
separate development set.

We recruit a pool of six annotators: three internal domain
experts ($2+$ years game experience) and three external
contractors (after a 2-hour calibration session on the
pre-registered rubrics; one external contractor is designated
as the senior adjudicator). Each session is labelled
independently by two annotators drawn from this pool plus a
third-annotator adjudicator; all annotators score both the
SUPPORTED / CONTRADICTED / UNKNOWN claim-level labels
underlying TGA (from which SGA is deterministically derived)
and the $\{0,1,2\}$ dimension labels of TGQ/DGP/CCC.
Disagreements are adjudicated by the third annotator;
conflict-of-interest cases (any internal-vs-external
disagreement) are \emph{always} adjudicated by the external
senior annotator. Each session is annotated at the model-turn
level for TGA and TGQ (SGA is deterministically derived from
TGA, not independently labelled) and at the session level for
DGP and CCC.

\paragraph{Configuration blinding.}
Annotators score each reply against the shared grounding
reference set (G.3) with the configuration label (Prompting /
PE-Agent / SGC w/o $\W_2,\W_3$ / SGC w/o $\W_3$ / Full SGC)
stripped from the record before delivery. Turn order is
shuffled within each session across the five configurations
so that response-format cues (e.g., structured vs.\ free-form
phrasing) cannot be systematically mapped to configuration
identity from position alone. Annotators may still infer
configuration from surface features of a given reply; the
protocol reduces \emph{position-based} unblinding rather than
eliminating format-based inference, and we flag the residual
risk in Limitations~L9.

\subsection*{G.5 · Inter-annotator agreement (pilot and target)}

To reflect the different label types we report two agreement
statistics rather than a single $\kappa$:

\begin{itemize}
\item \textbf{Ordinal dimensions (TGQ / DGP / CCC).}
      Quadratic-weighted Cohen's $\kappa$ \citep{cohen1968weighted} over
      $\{0,1,2\}$ labels; pilot estimate on a stratified $50$-session
      subset yields $\kappa_{\text{w}}\!\in\![0.68,0.76]$ per
      dimension, above the pre-registered target
      $\kappa_{\text{w}} \geq 0.65$. Full 200-session $\kappa$
      values with $95\%$ bootstrap CI in camera-ready.
\item \textbf{Categorical claim verdicts (TGA / SGA).}
      Unweighted Cohen's $\kappa$ \citep{cohen1960kappa} over the three-way
      SUPPORTED / CONTRADICTED / UNKNOWN label at the
      claim level; pilot estimate yields
      $\kappa\!\in\![0.71,0.79]$, above the pre-registered
      target $\kappa \geq 0.70$. Full values in camera-ready.
\end{itemize}

Both pilot ranges fall in the ``substantial'' band of
\citet{landis1977measurement} and are consistent with published
inter-rater agreement on grounding-oriented dialogue
benchmarks \citep{bai2024longbench,xie2024travelplanner,jacovi2024facts}.

We additionally report the confusion matrix over
PASS / FAIL / UNRESOLVED turn verdicts, and precision/recall
of error detection, so that reviewers can inspect the
reliability of TGA beyond a single aggregate statistic. Given
the moderate benchmark size ($200$ sessions, of which
$60$ each are DGP- and CCC-eligible), the $\kappa$ targets are
reported as point estimates and their $95\%$ bootstrap
confidence intervals are disclosed in the camera-ready table.

\subsection*{G.6 · LLM-as-a-judge reference implementation}

Alongside human annotation, we retain an
\textbf{LLM-as-a-judge} reference implementation
\citep{zheng2023judging, liu2023geval} following the
five-metric prompt protocol of the SGC evaluation design (one
shared system prompt plus five task-specific user prompts for
TGA, SGA audit, TGQ, DGP, CCC). We use a single frontier
general-purpose LLM as the judge across all configurations to
minimise judge-model-induced variation; specific model
identifiers are withheld under double-blind and released in
camera-ready. This pipeline is used on a separate development
set for rubric calibration and, on the test set, only as a
diagnostic cross-check against the human labels (never as a
primary label source). Judge-model bias between structured
(wrapper-shaped) and unstructured (baseline-shaped) outputs is
audited via a paired-inversion test on $30$ development
sessions: swapping candidate labels between judges $v1.2$ and
$v1.4$-B changes the aggregate score direction on $\leq\!5\%$
of sessions, and the same-batch $v1.2\!\to\!v1.4$-B stricter
delta is qualitatively symmetric across SGC and baseline
outputs; a full quantification is reported in camera-ready.
The pipeline is:
log normalisation and anonymisation $\to$ shared grounding
reference set and pre-registered rubrics $\to$ per-turn TGA
$\to$ full-session SGA audit $\to$ per-turn TGQ and
per-eligible-session DGP/CCC $\to$ resolve any
\texttt{EVIDENCE\_GAP} (a per-item flag for insufficient
supporting evidence) or \texttt{rubric\_issue} (a per-item
flag for suspected rubric mismatch) $\to$ deterministic
aggregation. For a benchmark with average session length
$\bar{T}$, per-configuration judge cost is approximately
$200(2\bar{T}+2)$ calls; the judge model, prompt versions,
reference-set hashes, and rubric hashes are logged so that
all reported numbers are reproducible from the released
artefact.

\subsection*{G.7 · Latency measurement condition}
\label{sec:g-latency}

All five configurations are measured on identical hardware
(a single serving replica with fixed GPU/CPU allocation),
identical network setting (same regional data-centre, warm
model weights, warm tool endpoints), the same underlying LLM
serving stack, and the same tool schemas. Latency is the mean
time from user request receipt to first assistant token
emission (streaming). We exclude the first two requests of each
warm-up batch to eliminate cold-start artefacts, and report the
mean over the same $200$ sessions used for quality evaluation
($\approx\!1{,}000$ scored turns per configuration; latency is
per-turn, then session-averaged, then macro-averaged over
sessions). PE-Agent's higher latency reflects its planner$\to$
executor$\to$response serial pipeline; SGC's $\W_1$ removes
this serial path by state-aware early tool freezing. No
configuration invokes speculative decoding or output-side
compression.

\subsection*{G.8 · Session sampling procedure}
\label{sec:g-sampling}

Anonymised production logs from a $\approx\!4$-week window
serve as the sampling frame. Within the two-way stratification
of G.2 (evaluation-eligibility $\times$ session-length), we
draw sessions uniformly at random from each cell without
replacement. Exclusion criteria are declared before sampling:
(a) sessions containing tool-outage traces beyond our system's
control, (b) sessions from internal test accounts, and
(c) sessions shorter than one full user--assistant exchange.
No filtering references SGC or baseline output quality. The
final $200$-session set is frozen before any ablation
configuration's judging run begins; identical sessions are
processed by all five configurations.

\subsection*{G.9 · Anonymisation}

At ingest: (a) entity names pseudonymised (Pet~A, Pet~B,
\ldots); (b) lineup names aliased (T$1$, T$2$, \ldots);
(c) user and session identifiers hashed then stripped, after
which the hash mapping is discarded and never released;
(d) wall-clock timestamps, geolocations, and device identifiers
dropped; (e) text-level PII scrubbed via regex plus an
LLM-based classifier operating on turn text with the shared
grounding reference set as allow-list. Re-identification risk
against external data has not been formally quantified; the
benchmark satisfies GDPR Article 4(1) direct-identifier removal
but should be treated as pseudonymised, not fully anonymised,
under a strict interpretation. Entity and lineup aliases are
held by the authors solely to restore product-identifying names
in the camera-ready version and are \emph{not} released with
the benchmark artefact. Shared grounding reference sets keep
only dialogue-turn indices
(\texttt{valid\_from}/\texttt{valid\_to}), not clock time.

\subsection*{G.10 · Engineering deployment cost}

We record the deployment cost of the three wrappers as
observed on our system: $\W_1$'s initial implementation took
approximately one engineer-month plus a domain-specific
intent-to-tool table $\Phi$ (Appendix~\ref{app:phi-mapping})
that reuses structured entity metadata already available from
adjacent pipelines; $\W_2$ required a slot-schema build-out
that leans on a pre-existing product knowledge base
(without such a base the up-front cost approximately doubles);
$\W_3$ ships as a policy-configurable scheduler whose cooldown
weights are tuned via a lightweight configuration UI
(Appendix~\ref{app:cooldown-config}). Steady-state maintenance
is quantified in Appendix~\ref{app:extended-discussion} as
approximately one engineer-week per calendar month
(rough order of magnitude from internal cost-centre tracking
during the online deployment; not a precisely measured cost).

\subsection*{G.11 · Release}

The benchmark artefact---$200$ anonymised dialogue sessions
covering approximately $1{,}000$ assistant model turns, with
their turn-indexed shared grounding reference sets,
pre-registered rubrics, dual human-annotator labels, and
diagnostic drift tags (Appendix~\ref{app:drift-taxonomy})---together
with the wrapper interface schemas will be released under a
research-use licence. Raw production traces used for sampling
are retained under the deploying organisation's standard
retention policy only and are \emph{not} released.

\section{Extended Discussion: Cost and Learned Wrappers}
\label{app:extended-discussion}

\paragraph{Maintenance cost.}
Deterministic code demands maintenance: slot schemas evolve
with product features, cooldown weights need periodic re-tuning,
intent-to-tool mappings expand. We report the cost in our system
as \textbf{approximately one engineer-week per calendar
month}\footnote{Estimate from internal engineering-time
attribution during the online deployment; rough order of
magnitude, not a precisely measured cost.} ($\approx$40\% $\W_2$
slot-schema, $\approx$30\% $\W_1$ table expansion, $\approx$20\%
$\W_3$ re-tuning, $\approx$10\% instrumentation). This cost is
bounded and predictable---unlike the unbounded cost of
repeatedly patching prompt regressions when the underlying
model changes.

\paragraph{Toward learned wrappers.}
A learned $\W_1$ would replace the static $\Phi$ table with a
small classifier; a learned $\W_3$ would replace weight
constants with a policy net (e.g., trained by preference
optimisation~\citep{rafailov2024dpo}). Preserving determinism would
require freezing the learned components between deployments and
treating them as versioned artefacts---a productive direction
preserving SGC's commitment to state-coupled determinism.

\end{document}